\documentclass[sigconf]{acmart}

\usepackage{booktabs} 
\usepackage [english]{babel}
\usepackage [autostyle, english = american]{csquotes}
\MakeOuterQuote{"}
\usepackage{subcaption}
\usepackage{url}
\usepackage{color}
\usepackage{array, caption, tabularx, makecell, booktabs}%
\usepackage{bold-extra}

\usepackage{multirow}

\setcopyright{rightsretained}
\acmConference{ICTIR' 17 Workshop on Search-Oriented Conversational AI (SCAI' 2017)}{October 1, 2017}{Amsterdam, Netherlands}

\begin{document}
\title[]{Combining Search with Structured Data to Create a More Engaging User Experience in Open Domain Dialogue}


\author{Kevin K. Bowden, Shereen Oraby, Jiaqi Wu, Amita Misra, and Marilyn Walker}
\affiliation{%
 \institution{Natural Language and Dialogue Systems Lab\\
 University of California, Santa Cruz}
}
\email{{kkbowden, soraby, jwu64, amisra2, mawalker}@ucsc.edu}

\renewcommand{\shortauthors}{K. K. Bowden et al.}




\begin{abstract}
The greatest challenges in building sophisticated open-domain
conversational agents arise directly from the potential for ongoing
mixed-initiative multi-turn dialogues, which do not follow a
particular plan or pursue a particular fixed information need. In
order to make coherent conversational contributions in this context, a
conversational agent must be able to track the types and attributes of
the entities under discussion in the conversation and know how they
are related.  In some cases, the agent can rely on structured
information sources to help identify the relevant semantic relations
and produce a turn, but in other cases, the only content available
comes from search, and it may be unclear which semantic relations hold
between the search results and the discourse context. A further
constraint is that the system must produce its contribution to the
ongoing conversation in real-time.  This paper describes our
experience building SlugBot for the 2017 Alexa Prize, and discusses
how we leveraged search and structured data from different sources to
help SlugBot produce dialogic turns and carry on conversations whose
length over the semi-finals user evaluation period averaged 8:17 minutes.
\end{abstract}

\maketitle

\section{Introduction}

The Alexa Prize funded 12 international teams to compete to create a
conversational agent that can discuss any topic for at least 20
minutes. UCSC's Slugbot was one of these funded teams.\footnote{The
  competition is still ongoing and its conditions prohibit us from
  reporting detailed information about the evaluation results or
  Slugbot's system architecture.} The greatest challenges with the
competition arise directly from the potential for ongoing
mixed-initiative multi-turn dialogues, which do not follow a
particular plan or pursue a particular fixed information need.  This
paper describes some of the lessons we learned building SlugBot for
the 2017 Alexa Prize, particularly focusing on the challenges of
integrating content found via search with content from structured data
in order to carry on an ongoing, coherent, open-domain,
mixed-initiative conversation. SlugBot's conversations over the semi-finals user evaluation averaged 8:17 minutes.

Unlike much previous work on conversational AI, SlugBot could not and
did not assume that the user had an ``information
need'' \cite{kiseleva2016predicting,chuklin2015click,Radlinski17}.
Rather, the design of the Alexa Prize was aimed at open conversations
that could engage the user, through any type of dialogue or chitchat,
discussing films and books, gossiping about celebrities, playing
verbal games, telling stories or sharing experiences, or any other of
many different types of activities that conversation is often used
for.

This open design foregrounds many longstanding challenges that
have not been solved even for task-oriented dialogue
systems. These include:

\begin{itemize}
\item Modeling discourse coherence;
\item Supporting mixed-initiative dialogue;
\item Generating contextualized and stylistically appropriate
natural language responses.
\end{itemize}

This paper is structured around the ``lessons learned'' with respect
to these challenges from our experience building SlugBot.  To be clear,
we are not offering a solution to these problems: instead our
intention is simply to highlight the difficulties with developing
adequate computational models of these phenomena that particularly
arise in the context of open-domain conversations, where users cannot
be assumed to be pursuing a particular task or 
information need.  We will attempt to motivate our hypothesis that a
comprehensive solution to these challenges for open-domain dialogue
requires a much deeper understanding and utilization of the semantic
relations that underly dialogue coherence.

\begin{table*}[ht]
\begin{small}
\centering
\begin{tabular}
{@{}p{0.3cm}|p{0.9cm}|p{6.8cm}|p{3in}@{}}
\toprule
S1 & System & {What do you want to talk about?} &  Opening prompt, expects to map an answer to a known topic.
\\ \hline
U2 & User & {Let's talk about movies.} & Response to open prompt, topic "movies"
\\ \hline
S3 & System & {I love movies! Which movies have you seen recently?} &
Take initiative to provide an opinion, then ask followup question. SlugBot
could have suggested talking about a particular movie, or asked if
the user had scene a particular movie.
\\ \hline
U4 & User & {I watched Jason Bourne recently.} & Names a particular film, named entity recognition must map ``Jason Bourne'' to a movie entity. 
\\ \hline
S5 & System & {Oh yes, according to ratings, Jason Bourne is a pretty good movie.$^{\ddagger}$} & SlugBot takes initiative, provide ratings information, not necessarily unique or what is asked for, conversational marker "Oh yes". Information provided from structured data from IMBD. 
\\ \hline
U6 & User & {I can't remember the actor's name in the movie, who stars in it?} &
Somewhat indirect question, conversational formulation "can't remember",
requires coreference resolution of "the movie" and "it" to "the movie named Jason Bourne" \\ \hline
S7 & System & {It stars Matt Damon.$^{\dagger (or \ddagger)}$} &
SlugBot provides a direct answer, recognizes that to maintain
discourse coherence it should use the pronoun "It". 
\\ \hline
U8 & User & {Have you heard much about it in terms of the plot?} &
Coreference resolution of "it", recognition that "the plot" is a related entity to a movie, and a valid conversational move. Need to map "the plot" 
to a known relation or search query.
\\ \hline
S9 & System & {The CIA's most dangerous former operative is drawn out of hiding to uncover more explosive truths about his past.$^{\ddagger}$} & System selects  first sentence of plot synopsis that is returned by search. Needs to know that the first sentence is usually good by itself.  \\ \hline
U10 & User & {Sounds about right. What other movies has Matt Damon been in?} & User requests topic chaining by actor to other movie entities. System must know that actor can be a key to find other movies that star that actor. \\
\bottomrule     
 \end{tabular}
\end{small}
\caption{Sample Dialogue about Movies. System content indicated
as coming  from search$^{\dagger}$ or structured data$^{\ddagger}$. 
 \label{table:sample-movie-dialogue}}
\vspace{-.7cm}
\end{table*}

For example, consider dialogue focused on content related to the movie
domain. This should be one of the easiest domains because it is
well-structured, and there are existing systems handling conversations
where there is a specified user information need or task, such as
finding films with particular properties, finding out what is playing
and where, or booking a movie ticket
\cite{kim2016natural,hakkani2014probabilistic,celikyilmaz2015enriching}. Moreover, the Internet Movie Database (IMDB) \cite{IMDb} provides information on
plot, rating, and actors that can be leveraged to support
conversations. IMDB also makes use of the Schema.org \cite{Schema}
structure to connect common entities to their related attribute types
(such as {\it Actor} $\rightarrow$ {\it Person} $\rightarrow$ {\it
  birthDate}), allowing the system to retrieve a large set of possible
next topics and related facts and entities.

However, remember that SlugBot is based on the assumption that the
user might simply enjoy talking about films and related entities and
therefore may freely move the conversational focus among different
movie entities, along with the vast array of semantically-associated
movie attributes: {\it movies} have actors, genres, plots, and awards;
{\it actors} have names, affiliations, other movies they were in,
awards, etc. {\it Actors} are people, who have spouses, families and
friends, and engage in other life activities besides acting, such as
political advocacy.

A potential dialogue is shown in
Table~\ref{table:sample-movie-dialogue}.\footnote{An example of a
  recently developed MovieBot that tries to support free-ranging
  conversations can be found at
  (\url{https://www.amazon.com/dp/B01MRKGF5W})} The interaction might
appear to be simple enough: the user chooses to discuss movies, and
selects {\it Jason Bourne} as the specific movie she is interested in,
the system finds the movie in IMDB, and then provides information on
its rating, lead actor, and plot. The user then changes the topic to
other movies with the same actor, and the conversation continues.

Even with the availability of IMDB, however, the interaction is not
totally straightforward. The RHS of
Table~\ref{table:sample-movie-dialogue} describes some of the required
competencies and decisions SlugBot must make. First, Slugbot must be
able to perform coreference resolution and recognize that {\it the
  movie} and {\it it} in turns U6 and U8 are coreferential.  We
estimate the accuracy of noun-phrase coreference resolution to only be
about 70\% for off-the-shelf tools applied to dialogue, since most of
them are targeted to text
\cite{Grosz77,Walker96b,BFP87,recasens2010coreference,PrinceWalker93,pradhan2011conll,recasens2013life,Manningetal14}.

More challenging is that at each system turn, there are a large number
of conversational moves that are possible.  Making good decisions
about what to say next requires balancing a dialogue policy as to what
dialogue acts might be good in this context, with real-time
information as to what types of content might be possible to use in
this context.  Slugbot could offer an opinion as in turn S3, ask a
follow-on question as in S3, take the initiative to provide unasked
for information, as in S5, or decide, e.g. in the case of the user's
request for plot information, to use search to retrieve some relevant
content. Search cannot be used effectively here without constructing
an appropriate query, or knowing in advance where plot information
might be available. In a real-time system, live search may not
be able to achieve the required speed and efficiency, so preprocessing
or caching of relevant information may be necessary. Finally,
most other domains do not have such high quality structured data available,
leaving us to develop or try to rely on more general models of discourse
coherence.

\section{Modeling Discourse Coherence}

In open-domain conversation, dialogue coherence between related turns
must be maintained.  What underlies dialogue coherence goes beyond
simple word overlap or similarity, and its clear that neural models of
open-domain conversational dialogue do not yet capture it.  Theories
of discourse posit that there are a small number of semantic relations
that can hold between adjacent turns: at the most general level these
are {\sc contingency}, {\sc comparison}, {\sc expansion}, and {\sc
  temporal order} \cite{Prasadetal08,Hobbs83,Hobbs90}.\footnote{Some
  theories rely on the plan-based coherence of discourse
  \cite{GS86,Litman85}, but we believe that this can only be easily
  applied to task-oriented dialogue.}  We posit that one way to allow
SlugBot to take the initiative and produce a turn that maintains
discourse coherence is to find content to use in Slugbot's next turn
that instantiates a valid semantic relation between the current user
turn and SlugBot's next turn.  One of the strongest bases for such
semantic relations are the relations captured by ontologies or frames,
which give us related entities, e.g. movies have actors and directors
\cite{hakkani2014probabilistic,heck2013leveraging}. These types of
relations can be used to instantiate the {\sc expansion} relation,
which basically captures moving to strongly related subtopics, often
by chaining off a particular discourse entity. To find content to
instantiate the {\sc expansion} relation to use in Slugbot's next turn
(taking the initiative), we carry out the following pipeline:

\begin{enumerate}
\item Perform coreference resolution on the user's turn
\item Identify relevant entities in the user query, i.e. a movie name or sports team, by parsing and checking entities using Google Knowledge Graph.
\item Look up information on the entity to assign relevant attributes, using tools like Schema.org, YAGO, or DBPedia (for example, a Sports Team has players).
\item Retrieve relevant information about the entity. In the easiest scenario, we might have a structured source of information, like IMBD. In other cases, we need to use search. This often then requires further processing, such as parsing unstructured information on Wikipedia or in search results and attempting
to extract relevant content.
\item Select content and produce an utterance to give back to the user. This may be accomplished through natural language generation , string selection, or sentence compression \cite{Walkeretal07,OrabyWalker17,Nayaketal17,Krauseetal17,Ericetal17}. We discuss NLG in more detail in Section~\ref{sec:nlg}.
\end{enumerate}

In the case of movies, the structure of IMDB, as discussed above,
allows us to link between related entities and attributes using
various DB keys.  However other conversational domains do not have
freely available richly structured information such as this.  It is
rare for a single resource to aggregate all the information that might
be useful, so SlugBot must be able to leverage information and
integrate information from multiple sources.  But state-of-the-art
knowledge bases and ontologies are still
limited. Table~\ref{table:resources} lists some of the resources that
we have found to be most useful for search and structured information.

\begin{table}[h]
\begin{small}
\centering
\begin{tabular}
{@{}p{0.1cm}|p{1.5cm}|p{6cm}@{}}
\toprule
\bf \# & \bf Tool & \bf Description \\
\midrule
1 & Wikipedia \cite{Wikipedia} & Multi-lingual, web-based, free-content encyclopedia.\\ \midrule

2 & YAGO \cite{Rebele2016} & Semantic knowledge base from Wikipedia \cite{Wikipedia}, WordNet \cite{WordNet}, and GeoNames \cite{GeoNames}, with over 10 million entities and over 120 million facts about them. \\ \midrule

3 & DBPedia \cite{DBPedia} & Crowd-sourced semantic knowledge graph using Wikipedia data (around 4.22 million entities in the ontology).\\ \midrule

4 & Google Knowledge Graph \cite{GoogleKnowledgeGraph} & API for finding entity information include type and details, with some relevance score.\\
\bottomrule     
 \end{tabular}
 \end{small}
 \caption{Search and Structured Information Resources}
\label{table:resources}

\vspace{-.7cm}
\end{table}

\begin{figure*}
  \centering
  \begin{subfigure}[b]{0.47\textwidth}
    \includegraphics[width=\textwidth]{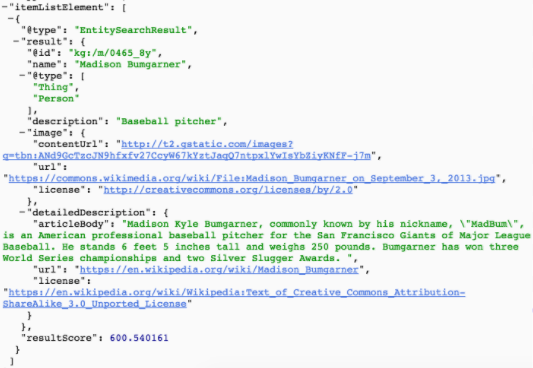}
    \caption{Google Knowledge Graph Search API Result}
    \label{fig:google}
  \end{subfigure}
  \hspace{2em}
  ~ 
  \begin{subfigure}[b]{0.3\textwidth}
    \includegraphics[width=\textwidth]{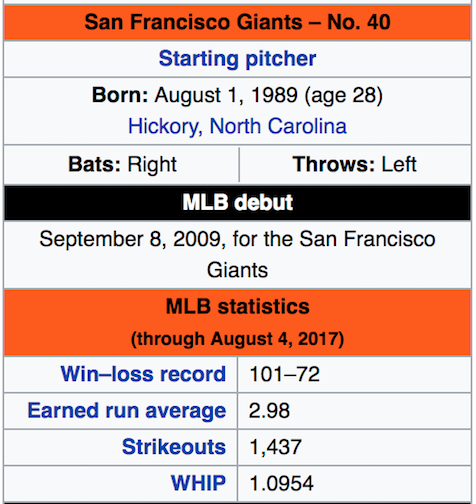}
    \caption{Wikipedia Result Table}
    \label{fig:wiki}
  \end{subfigure}
  
  ~ 
  \caption{Sample Available Resources for Query "Madison Bumgarner"}\label{fig:resource-figures}
\end{figure*}


Like movies, sports is another domain that has rich structure, and in
which there is broad user interest. Search results for a query
about "Madison Bumgarner" are in Figure~\ref{fig:resource-figures},
showcasing a sample of the different information retrievable from each source
(Step~2 of the pipeline). 

From the Google Knowledge Graph (Figure~\ref{fig:google} result we are able to ascertain the entity type, a brief description, and a relevant Wikipedia page (Figure~\ref{fig:wiki}) which we can use to find accurate structured information. We may further augment our knowledge by using the information returned by the Google Knowledge Graph as parameters to our YAGO or DBpedia query which can more easily extract specific relationships between
an entity-attribute. For example, the results returned by YAGO for the "Madison Bumgarner" query contains a connection to the headline \textit{Struggling MadBum might not garner next start}, which is contextually relevant data not encapsulated anywhere in the previously examined results.

There, however, exists a disconnect between the resources, i.e. some entities are available in one resource and not
another, or there may be inconsistent information across resources.
While it would be nice not to have to anticipate the types of
integration that are needed, our take-away from this, is that at
present, it appears we have to accomplish the steps in our pipeline by
integrating knowledge from different resources in advance, even though
projects such as YAGO have already been
working on such integration for at least ten years.

Other discourse coherence relations besides {\sc expansion} are also
viable candidates for selecting content for next turns, but finding
content that instantiates these relations can be a challenging problem
in itself. For example, in casual conversation, it is common to
provide opinions and then perhaps further take the initiative and
justify them. The justification of an opinion is a type of {\sc
  contingency} relation: we describe how we curate content to provide
justifications in Section \ref{sec:mixinit}.

We have also been able to use the {\sc temporal} relation in a limited
way by drawing on narratively structured sources, such as personal
stories in blogs. Since these stories are told in temporal order, we
can repurpose the content of these blogs to tell stories, maintaining
pre-existing narrative coherence when the system produces a sequence
of turns \cite{Bowdenetal16}. However, we posit that there is much
more that could be done to make better use of deep semantic discourse
relations for recognizing discourse relations and generating coherent
conversational turns.

\section{Mixed Initiative Dialogue}
\label{sec:mixinit}

Mixed Initiative dialogue is key to a natural conversational
interaction
\cite{HallerMcroy98,Chu-Carroll98,WW90,Allen99,Bohus03,Radlinski17},
and this is even more important for open domain dialogue than it is
for task-oriented or information seeking dialogue.  One of our primary
hypotheses, as described above, is that good models of discourse
coherence will help SlugBot identify content that can be used to take
the initiative. However, models of discourse coherence have been
rarely applied to conversation
\cite{stent2000rhetorical,riccardi2016discourse,tonelli2010annotation}
and thus there is considerable work to be done simply in understanding
how these relations can be instantiated in dialogue.

In addition, a further challenge arises from the fact that both system
and user options for {\bf dialogue acts} are extremely varied at each
turn, e.g. user intents can be to provide opinions, give or solicit
information, contrast two possibilities, request the system to perform
an action, and more. One reasonable taxonomy for the types of dialogue
acts that might be available to SlugBot could be based for example on
the dialogue act annotations in the Switchboard
corpus \cite{SWBD-DAMSL}.

\begin{table}[ht]
\begin{small}
\centering
\begin{tabular}
{p{1cm}|p{1cm}|p{1.2cm}|p{2.8cm}}
\toprule
Entity & Bin & Sentiment & Justifications\\
\hline
Magneto & comics & 5 & he can control metal\\
Aliens & movie & 4 & well cast,action packed\\
Dracula & monster & 5 & scary\\
\bottomrule     
 \end{tabular}
 \caption{Example opinion data used in Table \ref{table:comics} and \ref{table:monsters}}\label{table:opinions_data}
\end{small}
\vspace{-.8cm}
\end{table}

\begin{table}[ht]
\begin{small}
\begin{tabular}
{@{}p{0.1cm}|p{0.9cm}|p{6.8cm}@{}}
\toprule
1 & System & {What do you want to talk about?} \\
2 & User & {Let's talk about comic books.} \\
3 & System & {I'm a pretty avid comic book reader! Who is your favorite character?} \\
4 & User & {I really like Moon Knight} \\
5 & System & {Moon Knight is a fictional superhero appearing in American comic books published by Marvel Comics. Created by Doug Moench and Don Perlin, the character first appeared in Werewolf by Night issue 32$^{\dagger}$. Does it sound like we are thinking of the same character? Why do you like them?} \\
6 & User & {Yep that's the one, I like him because he's edgy. Who is your favorite character?} \\
7 & System & {I like Magneto.$^{\ddagger}$} \\
8 & User & {Okay why?} \\
9 & System & {Personally, I think he's cool because he can control metal.$^{\ddagger}$} \\
10 & User & {I guess that is a cool power, when did he first appear?} \\
11 & System & {I think that Magneto first appeared in X-men issue 1 in 1963.$^{\ddagger (or \dagger)}$} \\
\bottomrule        
 \end{tabular}
\end{small}
\caption{Sample Dialogue about Comic Books. System content  based on
either search$^{\dagger}$ or structured data$^{\ddagger}$. \label{table:comics}}
\vspace{-.3in}
\end{table}

Here, we consider a simple case combining discourse relations and
dialogue acts that we have implemented in Slugbot in order to take the
initiative in a way that we hoped the user would find interesting.
Our aim was to utilize the {\sc contingency} discourse relation to
connect a statement of opinion and its justification. We designed a
template containing both arguments of the {\sc contingency} relation,
namely \textit{I think} $\{entity\}$ \textit{is} $\{sentiment\}$
\textit{because} $\{justification\}$.  We construct a table of
argument pairs that can instantiate this relation, as shown in Table
\ref{table:opinions_data}.  This table can be populated by
crowd-sourcing or by using search as a pre-processing step.

Table \ref{table:comics} illustrates how this is used in our
conversations about comics. At Line 6, when the user asks \textit{Who
  is your favorite character?}, it is most appropriate to provide an
opinion. It is difficult to imagine retrieving search-based data which
contains a contextually relevant opinion, but it is even more
difficult to imagine that if search had returned such an opinion, that
search could be used a second time in order to retrieve a
justification for the provided opinion and answer the user's follow-up
question in Line 8, {\it Okay why?}. The source text for the search
would have to be annotated for the type of content that could be used
to provide justifications, and search would have to support these
types of semantic relations.

\section{Natural Language Generation}
\label{sec:nlg}

The current challenges for natural language generation, in our view,
arise from the need to combine information from structured and
unstructured sources when producing conversational utterances. SlugBot
currently uses a combination of pre-written templates, sentence
selection, and techniques for telling stories that are based
on converting monologic stories to dialogic sequences \cite{Bowdenetal16}.

Structured data, when available, can do more than structure a search
result: it can also be easier to use within a conversation because it
provides the necessary structure needed for high precision natural language generation
\cite{Walkeretal07,StentMolina09}. More precisely, a small set of
generic templates with various slots can be filled with information
from structured data sources to insure high quality, accurate
responses. These generic templates can be hand crafted, or prepared in
advance by learning natural language generation templates
automatically from appropriate conversational domain sources such as
different types of user-generated content
\cite{Higashinakaetal07,OrabyWalker17}, as illustrated in our justification
initiatives above in Section~\ref{sec:mixinit}.

For general fact-based questions, on the other hand, search content
can be used directly.  For example, at line 14 in Table
\ref{table:monsters} when the user asks \textit{What was the first
  movie to feature a vampire?}, search provides us with a good
response. This introduces however the challenge of updating the
discourse context with the right representation of the two movies
under discussion, so that they can then be available for follow-on
coreference.  This is an open problem.

\begin{table}
\begin{small}
\begin{tabular}
{@{}p{0.1cm}|p{0.9cm}|p{6.8cm}@{}}
\toprule
1 & System & {You know a secret passion of mine is talking about famous monsters from monster movies. Want to talk about that?} \\
2 & User & {Sure.} \\
3 & System & {Great, do you have a specific type of monster movie that you like?} \\
4 & User & {I guess, I mean I like movies about aliens.} \\
5 & System & {Did you have a specific alien movie in mind?$^{\ddagger}$} \\
6 & User & {I just saw Aliens the other day. Can you tell me about it.} \\
7 & System & {Aliens is a 1986 American science-fiction action horror film written and directed by James Cameron and starring Sigourney Weaver It is the sequel to the 1979 film Alien and the second installment in the Alien franchise.$^{\dagger}$} \\
8 & User & {What did you think of it?}\\
9 & System & {I though Aliens was well cast and is action packed.$^{\ddagger}$}\\
10 & User & {Same} \\
11 & System & {Do you want to hear about my favorite monsters?} \\
12 & User & {Absolutely} \\
13 & System & {I like Dracula from the Dracula movies. Have you heard of him?$^{\ddagger}$}\\
14 & User & {What was the first movie to feature a vampire?}\\
15 & System & {Nosferatu is the first film to feature a Vampire's death by sunlight, which formerly only weakened vampires. The next classic treatment of the vampire legend was in Universal's Dracula starring Bela Lugosi as Count Dracula.$^{\dagger}$}\\
\bottomrule        
 \end{tabular}
\end{small}
 \caption{Sample Dialogue about Monsters \label{table:monsters}.
System content is curated based on
search$^{\dagger}$ or structured data$^{\ddagger}$.}
\vspace{-.3in}
\end{table}

\begin{table}
\begin{small}
\begin{tabular}
{@{}p{0.1cm}|p{0.9cm}|p{6.8cm}@{}}
\multicolumn{3}{l}{\bf Structured Data} \\
\toprule
1 & User & {What do you know about the Hitchhiker's Guide to the Galaxy?} \\
2 & System & {The Hitchhiker's Guide to the Galaxy is a science fiction book from 1981.$^{\ddagger}$} \\
\hline
\multicolumn{3}{l}{\bf Search} \\
\hline
1 & User & {What do you know about the Hitchhiker's Guide to the Galaxy?} \\
2 & System & {The Hitchhiker's Guide to the Galaxy is the first of five books in the Hitchhiker's Guide to the Galaxy comedy science fiction "trilogy" by Douglas Adams.$^{\dagger}$} \\
\bottomrule        
 \end{tabular}
 \caption{Using Structured Data vs Search \label{table:search_vs_struct}}
\end{small}
\vspace{-.3in}
\end{table}

It is clear that in order to use a semi-structured approach, we need
to determine when to utilize each source.  Structured data can be
easier to formulate into system responses and can often more easily
handle on-topic follow-up questions, but is more limited in scope. An
obvious approach, also used in the Watson Jeopardy system
\cite{ferrucci2010building}, is to pool responses from both sources
and rank them. We have not, to date, collected enough data to
build a ranker.

Our plan is to apply a combination of reinforcement learning and
learning of ranking functions for utterance variants in a particular
context to SlugBot conversations as we move forward with our own data
collection, outside of the Alexa Prize competition
\cite{stanfordCoherence16,yu16ticktockRL,Walker00,WKL00,WBK99}.  The
first step however is to use the Alexa Prize competition data to learn
a Paradise-Open-Domain evaluation function, with additional metrics
relevant to open-domain dialogue, e.g. independent variable metrics
that predict overall dialogue quality such as response delay,
vocabulary diversity, dialogue act sequence n-grams
\cite{WalkerPassonneau01}, conversational depth, number of reprompts
\cite{WLHWG02}, and other measures that can be automatically
logged. Many of the required measures have been used over the last 20
years in Paradise to evaluate task-oriented dialogue systems and they
remain highly relevant to overall dialogue quality in open-domain
dialogue systems \cite{Walkeretal02b,WPB01,WLKA97}. We predict this
can potentially improve the overall performance of the system as
demonstrated in Table \ref{table:search_vs_struct}. Here, the
structured data is sparse, resulting in an uninteresting response,
while search returns a very robust answer. Our Paradise-Open-Domain
evaluation function would need to learn to place priority on the
result returned by search, through ranking, despite having structured
data.

For open domain NLG, we have also conducted experiments with neural
sequence to sequence approaches using open domain corpora such as film
dialogue, Big Bang theory scripts, and open subtitles. These
approaches to date do not produce interesting utterances that maintain
discourse coherence. It is possible that further curation and semantic
annotation of these resources, e.g. by labelling semantic roles and
identifying dialogue acts and discourse relations might be helpful,
but this could also introduce data sparsity. For example in
Switchboard the dialogue act distribution is highly skewed.
Integrating information across multiple sources could also be further
explored \cite{Bowdenetal16}. Recent work on hybrid neural generation
approaches that use knowledge of sentence and discourse planning
structures also seem promising
\cite{Nayaketal17,Walker00,Stentetal02}.

\section{Conclusions}
In this paper, we describe some of the challenges we encountered
building SlugBot, an open domain conversational agent funded by the
Amazon Alexa Prize. We have introduced more problems than we have
solved, and we have attempted to support our hypothesis that we need
richer models of discourse coherence and discourse semantics to allow
a conversational agent to take the initiative in open domain
conversations. We illustrated how search and structured information
can be combined in order for SlugBot to find content to use to take
the initiative and respond to the user's utterances. We propose a
hybrid approach for language generation that which combines templates
to generate responses with sentence selection from search, and we show
examples in different domains to demonstrate real-world use cases that
make use of our approach. For future work, we plan to bring together
resources that provide structured data from different sources into a
single, accessible framework, to supply personal assistants with
scalable knowledge bases that will power more natural, mixed
initiative, and engaging conversations. We believe that it will be
possible in the next few years to build conversational agents that can
carry on a conversation for 20 minutes about many different topics.

\bibliographystyle{plain}
\bibliography{scai-refs,nl,phd} 

\end{document}